\documentclass[11pt,a4paper]{article}
\usepackage[hyperref]{emnlp2018}
\usepackage{times}
\usepackage{latexsym}

\usepackage{times}
\usepackage{latexsym}
\usepackage{url}
\usepackage{helvet}  
\usepackage{courier}  
\usepackage{url}  
\usepackage{graphicx}  
\usepackage{algorithm}
\usepackage[noend]{algpseudocode}
\usepackage{eufrak}
\usepackage{amsmath}
\usepackage{pgfplots}
\usepackage{tikz}
\usepackage{multirow}
\include{plots}
\definecolor{bblue}{HTML}{4F81BD}
\definecolor{rred}{HTML}{C0504D}
\definecolor{ggreen}{HTML}{9BBB59}
\definecolor{ppurple}{HTML}{9F4C7C}
\definecolor{Dark scarlet}{HTML}{560319}
\definecolor{Forest green}{HTML}{1E4D2B}

\usepackage{url}
\usepackage{color}
\usepackage{multirow}
\usepackage{amsfonts}
\usepackage{float}
\usepackage[export]{adjustbox}
\usepackage{adjustbox}
\usepackage{booktabs}
\usepackage{array}

\usepackage{eufrak}

\usepackage{amsmath}

\usepackage{algorithm}
\usepackage[noend]{algpseudocode}

\makeatletter
\def\BState{\State\hskip-\ALG@thistlm}
\makeatother

\usepackage{graphicx}
\usepackage{caption}
\usepackage{subcaption}

\aclfinalcopy 

\usepackage{mathtools}
\DeclarePairedDelimiter\floor{\lfloor}{\rfloor}

\title{
LISA: Explaining Recurrent Neural Network Judgments via Layer-wIse Semantic Accumulation and Example to Pattern Transformation}

\newcommand*{\affaddr}[1]{#1} 
\newcommand*{\affmark}[1][*]{\textsuperscript{#1}}

\author{Pankaj Gupta\affmark[1,2], Hinrich Sch\"{u}tze\affmark[2]\\ 
 \affaddr{\affmark[1]Corporate Technology, Machine-Intelligence (MIC-DE), Siemens AG  Munich, Germany}\\
  \affaddr{\affmark[2]CIS, University of Munich (LMU) Munich, Germany} \\
  {\tt pankaj.gupta@siemens.com}\\
  {\tt pankaj.gupta@campus.lmu.de |  inquiries@cislmu.org}
}

\date{}

\begin{document}
\maketitle
\begin{abstract}
Recurrent neural networks (RNNs) are temporal networks and cumulative in nature that have shown promising results  
in various natural language processing tasks. Despite their success, it still remains a challenge to understand their hidden behavior.   
In this work, we analyze and interpret the cumulative nature of  RNN via a proposed technique named as {\it Layer-wIse-Semantic-Accumulation} (LISA)  
for explaining decisions and 
detecting the most likely (i.e., saliency) patterns that the network relies on while decision making.  
We demonstrate (1) {\it LISA}: ``{How an RNN accumulates or builds semantics during its sequential processing for a given text example and expected response}"
 (2) {\it Example2pattern}: 
``{How the saliency patterns look like for each category in the data according to the network in decision making}".  
We analyse the sensitiveness of RNNs about different inputs to check the increase or decrease in prediction scores and further extract 
the saliency patterns learned by the network. 
We employ two relation classification datasets: SemEval 10 Task 8 and TAC KBP Slot Filling to 
explain RNN predictions via the {\it LISA} and  {\it example2pattern}. 

\end{abstract}

\section{Introduction}
The interpretability of systems based on deep neural network is required to be able to explain the reasoning behind the network prediction(s), 
that offers to (1) verify that the network works as expected and identify the cause of incorrect decision(s) 
(2) understand the network in order to improve data or model with or without human intervention.  
There is a long line of research in techniques of interpretability of Deep Neural networks (DNNs) via different aspects, such as explaining network decisions, data generation, etc.  
\newcite{erhan2009visualizing, hinton2012practical, simonyan2013deep} and \newcite{nguyen2016synthesizing} focused on model aspects to interpret neural networks via 
activation maximization approach by finding inputs that maximize activations of given neurons. 
\newcite{goodfellow6572explaining} interprets by generating adversarial examples.  
However, \newcite{baehrens2010explain} and \newcite{bach2015pixel, montavon2017methods} explain neural network predictions by sensitivity analysis to different input features 
and decomposition of decision functions, respectively. 

\begin{figure*}[t]
{
  \centering
   \includegraphics[scale=0.5]{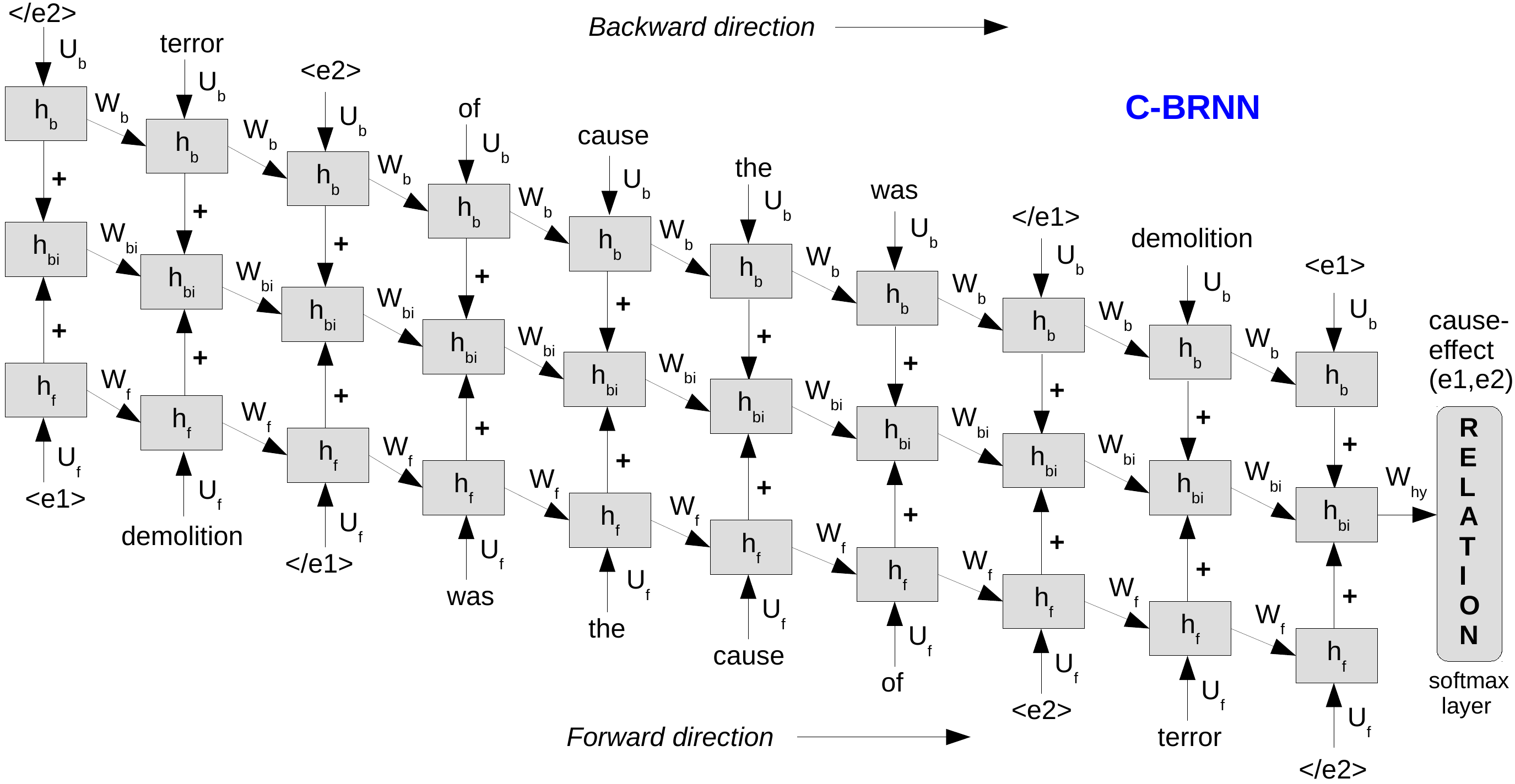}
    \caption{Connectionist Bi-directional Recurrent Neural Network (C-BRNN) \cite{Thang:82}}
    \label{fig:CBRNN}
}
\end{figure*}

\begin{figure*}[t]
{
  \centering
   \includegraphics[scale=0.38]{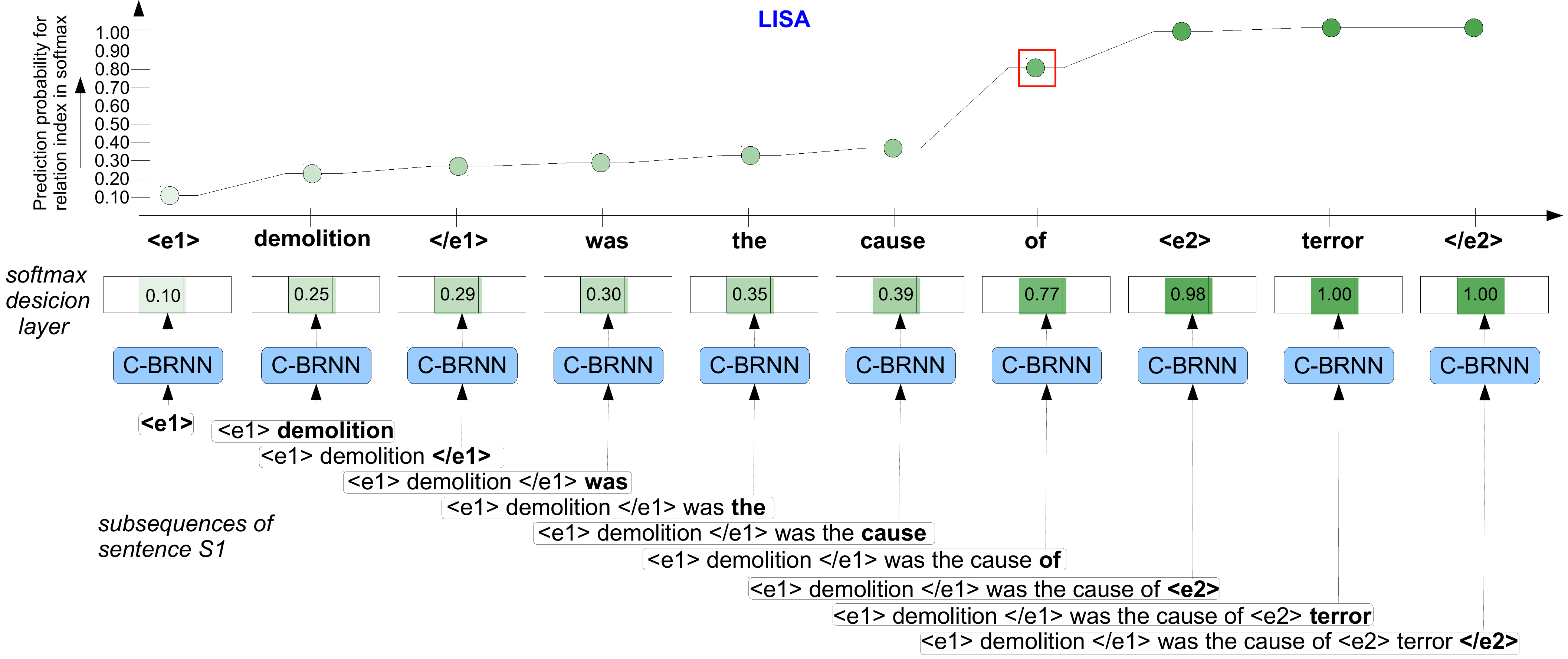}
    \caption{An illustration of Layer-wIse Semantic Accumulation (LISA) in C-BRNN, where we compute prediction score for a (known) relation type at each of the input subsequence.  
The highlighted indices in the softmax layer signify one of the relation types, i.e., {\it cause}-{effect}(e1, e2) in SemEval10 Task 8 dataset. The bold signifies the last word in the  subsequence.  
Note: Each word is represented by N-gram (N=3, 5 or 7), therefore each input subsequence is a sequence of N-grams. E.g., the word `of' $\rightarrow$ `cause of $<$e2$>$' for N=3.  
To avoid complexity in this illustration, each word is shown as a uni-gram.} 
    \label{fig:scorecomputing}
}
\end{figure*}

Recurrent neural networks (RNNs) \cite{elman1990finding} are temporal networks and cumulative in nature to effectively model sequential data such as 
text or speech.  RNNs and their variants such as LSTM \cite{hochreiter1997long} have shown success in several natural language processing (NLP) tasks, such as 
entity extraction \cite{LampleLSTMCRF, EduardLSTMCNNsCRF}, relation extraction \cite{Thang:82, Miwabansal2016,GuptaTFMTRNN2016,GuptaCrossRE:82}, language modeling \cite{mikolov2010recurrent, PetersELMO2018}, slot filling \cite{mesnil2015using, Thang:81}, machine translation \cite{bahdanau2014neural}, sentiment analysis \cite{ZhaoSentimentLSTM,TingSentimentRNN}, semantic textual similarity \cite{mueller2016siamese, gupta2018replicated}
and dynamic topic modeling \cite{GuptaRNNRSM:82}. 

Past works \cite{zeiler2014visualizing,dosovitskiy2016inverting} have mostly analyzed deep neural network, especially CNN in the field of computer vision 
to study and visualize the features learned by neurons.  Recent studies have investigated visualization of RNN and its variants.   
\newcite{tang2017memory} visualized the memory vectors to understand the behavior of LSTM and gated recurrent unit (GRU) in speech recognition task. 
For given words in a sentence, \newcite{LIsensitivityRNN2016} employed heat maps to study sensitivity and meaning composition in recurrent networks.  
\newcite{ming2017understanding} proposed a tool, RNNVis to visualize hidden states based on RNN's expected response to inputs.  
\newcite{PetersELMO2018} studied the internal states of deep bidirectional language model to learn contextualized word representations and  
observed that the higher-level hidden states capture word semantics, while lower-level states capture syntactical aspects. 
Despite the possibility of visualizing hidden state activations and performance-based analysis, there still remains a challenge for humans to interpret hidden behavior of the``black box" networks 
that  raised questions in the NLP community  as to verify  that the network behaves as expected. 
In this aspect, we address the cumulative nature of RNN with the text input and computed response to answer  
``how does it aggregate and build the semantic meaning of a sentence word by word at each time point in the sequence for each category in the data". 

{\bf Contribution}: In this work, we analyze and interpret the cumulative nature of  RNN via a proposed technique named as {\it Layer-wIse-Semantic-Accumulation} (LISA)  
for explaining decisions and 
detecting the most likely (i.e., saliency) patterns that the network relies on while decision making.  
We demonstrate (1) {\it LISA}: ``{How an RNN accumulates or builds semantics during its sequential processing for a given text example and expected response}"
 (2) {\it Example2pattern}: 
``{How the saliency patterns look  like for each category in the data according to the network in decision making}".  
We analyse the sensitiveness of RNNs about different inputs to check the increase or decrease in prediction scores. 
For an example sentence that is classified correctly, we identify and extract a saliency pattern  (N-grams of words in order learned by the network) 
that contributes the most in prediction score.  Therefore, the term {\it example2pattern} transformation for each category in the data.  
We employ two relation classification datasets: SemEval 10 Task 8 and TAC KBP Slot Filling (SF) Shared Task (ST) to 
explain RNN predictions via the proposed {\it LISA} and  {\it example2pattern} techniques.

\section{Connectionist Bi-directional RNN} 
We adopt the bi-directional recurrent neural network architecture with ranking loss, proposed by \newcite{Thang:82}. 
The network consists of three parts: 
a forward pass which processes the original sentence word by word (Equation \ref{eq:forward}); 
a backward pass which processes the reversed sentence word by word (Equation \ref{eq:backward}); 
and a combination of both (Equation \ref{eq:combined}). The forward and backward passes are combined by adding their hidden layers. 
There is also a connection to the previous combined hidden layer with weight  $W_{bi}$ with a motivation to 
include all intermediate hidden layers into the final decision of the network (see Equation \ref{eq:combined}). 
They named the neural architecture as `Connectionist Bi-directional RNN' (C-BRNN). 
Figure \ref{fig:CBRNN} shows the C-BRNN architecture, where all the three parts are trained jointly. 
\begin{align}
\begin{split}\label{eq:forward}
h_{f_t} = f(U_f \cdot w_t + W_f \cdot h_{f_{t-1}})
\end{split}\\
\begin{split}\label{eq:backward}
h_{b_t} = f(U_b \cdot w_{n-t+1} + W_b \cdot h_{b_{t+1}})
\end{split}\\
\begin{split}\label{eq:combined}
h_{{bi}_t} = f(h_{f_t} + h_{b_t} + W_{bi} \cdot h_{{bi}_{t-1}})
\end{split}
\end{align}

where $w_t$ is the word vector of dimension $d$ for a word at time step $t$ in a sentence of length $n$. $D$ is the hidden unit dimension.
$U_f \in \mathbb{R}^{d \times D}$ and $U_b \in \mathbb{R}^{d \times D}$  are the weight matrices between hidden units and input $w_t$ in forward and backward networks, respectively;
$W_f  \in \mathbb{R}^{D \times D}$ and $W_b  \in \mathbb{R}^{D \times D}$ are the weights matrices connecting hidden units in forward and backward networks, respectively. 
$W_{bi}  \in \mathbb{R}^{D \times D}$ is the weight matrix connecting the hidden vectors of the combined forward and backward network. 
Following \newcite{Gupta:89} 
during model training, we use 3-gram and 5-gram representation of each word $w_t$ at timestep $t$ in the word sequence, 
where a 3-gram for $w_t$ is obtained by concatenating the corresponding word embeddings, i.e., $w_{t-1} w_t w_{t+1}$. 

{\bf Ranking Objective}:
Similar to \newcite{Santos:82} and \newcite{Thang:82}, we applied the ranking loss function to train C-BRNN. 
The ranking scheme offers to maximize the distance between the true label $y^{+}$ and the best competitive label $c^{-}$ given a data point $x$. 
It is defined as-
\begin{align}
\begin{split}
{\cal L} = \log (1 + \exp(\gamma(m^+ -  s_{\theta} (x)_{y^+} )))\\ + \log(1 + \exp( \gamma (m^- + s_{\theta} (x)_{c^-}))) 
\end{split}
\end{align}

where $s_{\theta} (x)_{y^+}$ and  $s_{\theta} (x)_{c^-}$ being the scores for the classes $y^+$ and 
$c^-$, respectively. The parameter $\gamma$ controls the penalization of the prediction errors and $m^+$ and $m^−$ are margins for the correct and incorrect classes. 
Following \newcite{Thang:82}, we set $\gamma$ = 2, $m^+$ = 2.5 and  $m^-$ = 0.5.

{\bf Model Training and Features}:
We represent each word by the concatenation of its word embedding and position feature vectors. 
We use word2vec \cite{Mikolov:82} embeddings, that are updated during model training.  
As position features in relation classification experiments, we use position indicators (PI) \cite{ZhangandWang:82} 
in C-BRNN to annotate target entity/nominals in the word sequence, 
without necessity to change the input vectors,
while it increases the length of the input word sequences, as four independent words, as
position indicators ($<$e1$>$, $<$/ e1$>$, $<$e2$>$, $<$/e2$>$) around the relation arguments are introduced.

In our analysis and interpretation of recurrent neural networks, we use the trained C-BRNN (Figure \ref{fig:CBRNN}) \cite{Thang:82} model.  

\section{LISA and Example2Pattern in RNN}\label{sec:LISAexample2pattern}
There are several aspects in interpreting the neural network, for instance via 
(1) {\it Data}: ``Which dimensions of the data are the most relevant for the task'' 
(2) {\it Prediction} or {\it Decision}: ``Explain why a certain 
pattern" 
is classified in a certain way 
(3) {\it Model}: ``How patterns belonging to each category in the data look like according to the network". 

In this work, we focus to explain RNN via {\it decision} and {\it model} aspects by finding the patterns that explains ``why" a model arrives at a particular decision for each category in the data and verifies that model behaves as expected.  
To do so, we propose a technique named as LISA that interprets RNN about 
``how it accumulates and builds meaningful semantics of a sentence word by word"  
and ``how the saliency patterns look like according to the network" for each category in the data while decision making.  
We extract the saliency patterns via {\it example2pattern} transformation. 

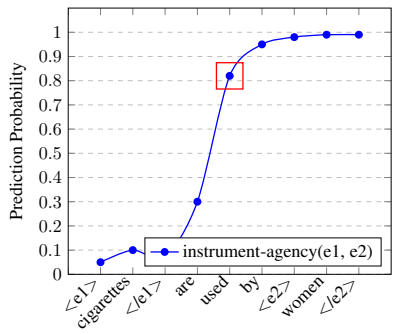
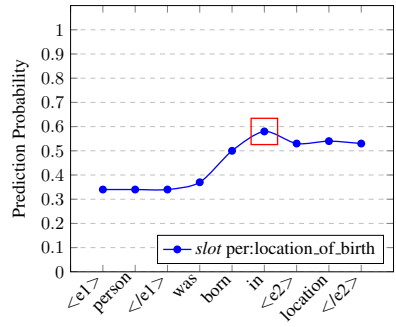
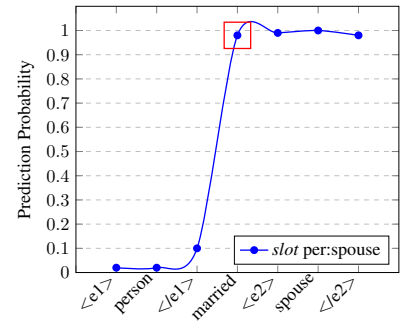
\begin{figure*}[] 
\begin{subfigure}{0.33\linewidth}
\centering
\begin{tikzpicture}[scale=0.62,trim axis left, trim axis right]
\begin{axis}[
    ylabel={Prediction Probability},
    xmin=0, xmax=11,
    ymin=0.0, ymax=1.1,
    xtick={0,1,2,3,4,5,6,7,8,9,10},
    ytick={0,0.1,0.2,0.3,0.4,0.5,0.6,0.7, 0.8, 0.9, 1.0},
    xticklabels={,$<$e1$>$, demolition,  $<$/e1$>$, was, the, cause, of, $<$e2$>$, terror, $<$/e2$>$},
    x tick label style={rotate=45,anchor=east},
    legend pos=south east,
    ymajorgrids=true,
    grid style=dashed,
   xlabel near ticks, compat=1.3
]
\addplot[thick, 
    color=blue,
    mark=*,
   mark options={solid}, smooth
    ]
    plot coordinates {
    (1,0.1)
    (2,0.25)
    (3,0.29)
    (4,0.30)
    (5,0.35)
    (6,0.39)
    (7,0.77)
    (8,0.98)
    (9,1.0)
    (10,1.0)
    };

\addplot[thick, 
    color=red,
    mark=square, mark size=8pt
    ]
    plot coordinates {
    (7,0.77)
    };

\addlegendentry{cause-effect(e1, e2)}
\end{axis}
\end{tikzpicture}%
\caption{{\it LISA} for $S1$}\label{LISAforS1}
\end{subfigure}\hspace*{\fill}%
~%
\begin{subfigure}{0.33\textwidth}
\centering
\begin{tikzpicture}[scale=0.62,trim axis left, trim axis right]
\begin{axis}[
    ylabel={Prediction Probability},
    xmin=0, xmax=10,
    ymin=0.0, ymax=1.1,
    xtick={0,1,2,3,4,5,6,7,8,9},
    ytick={0,0.1,0.2,0.3,0.4,0.5,0.6,0.7, 0.8, 0.9, 1.0},
    xticklabels={,$<$e1$>$, damage,  $<$/e1$>$, caused, by, the, $<$e2$>$, bombing, $<$/e2$>$},
    x tick label style={rotate=45,anchor=east},
    legend pos=south east,
    ymajorgrids=true,
    grid style=dashed,
 xlabel near ticks, compat=1.3
]
\addplot[thick, 
    color=blue,
    mark=*,
   mark options={solid}, smooth
    ]
    plot coordinates {
    (1,0.04)
    (2,0.18)
    (3,0.74)
    (4,0.76)
    (5,0.98)
    (6, 1.0)
    (7,1.0)
    (8,1.0)
    (9,1.0)
    };

\addplot[thick, 
    color=red,
    mark=square, mark size=8pt
    ]
    plot coordinates {
     (3,0.74)
    };

\addlegendentry{cause-effect(e2, e1)}
\end{axis}
\end{tikzpicture}%
\caption{{\it LISA} for $S2$}\label{LISAforS2}
\end{subfigure}\hspace*{\fill}%
~%
\begin{subfigure}{0.33\textwidth}
\centering
\begin{tikzpicture}[scale=0.62,trim axis left, trim axis right]
\begin{axis}[
    ylabel={Prediction Probability},
    xmin=0, xmax=9,
    ymin=0.0, ymax=1.1,
    xtick={0,1,2,3,4,5,6,7,8},
    ytick={0,0.1,0.2,0.3,0.4,0.5,0.6,0.7, 0.8, 0.9, 1.0},
    xticklabels={,$<$e1$>$, courtyard,  $<$/e1$>$, of, the, $<$e2$>$, castle, $<$/e2$>$},
    x tick label style={rotate=45,anchor=east},
    legend pos=south east,
    ymajorgrids=true,
    grid style=dashed,
     xlabel near ticks, compat=1.3
]
\addplot[thick, 
    color=blue,
    mark=*,
   mark options={solid}, smooth
    ]
    plot coordinates {
    (1,0.23)
    (2,0.25)
    (3,0.19)
    (4,0.98)
    (5,0.99)
    (6,0.99)
    (7,1.0)
    (8,1.0)
    };

\addplot[thick, 
    color=red,
    mark=square, mark size=8pt
    ]
    plot coordinates {
    (4,0.98)
    };

\addlegendentry{component-whole(e1, e2)}
\end{axis}
\end{tikzpicture}%
\caption{{\it LISA} for $S3$}\label{LISAforS3}
\end{subfigure}

\medskip

\begin{subfigure}{0.33\textwidth}
\centering
\begin{tikzpicture}[scale=0.62,trim axis left, trim axis right]
\begin{axis}[
    ylabel={Prediction Probability},
    xmin=0, xmax=11,
    ymin=0.0, ymax=1.1,
    xtick={0,1,2,3,4,5,6,7,8,9,10},
    ytick={0,0.1,0.2,0.3,0.4,0.5,0.6,0.7, 0.8, 0.9, 1.0},
    xticklabels={,$<$e1$>$, marble,  $<$/e1$>$, was, dropped, into, the, $<$e2$>$, bowl, $<$/e2$>$},
    x tick label style={rotate=35,anchor=east},
    legend pos=south east,
    ymajorgrids=true,
    grid style=dashed,
    xlabel near ticks, compat=1.3
]
\addplot[thick, 
    color=blue,
    mark=*,
   mark options={solid}, smooth
    ]
    plot coordinates {
    (1,0.11)
    (2,0.24)
    (3,0.28)
    (4,0.39)
    (5,0.39)
    (6,0.89)
    (7,0.99)
    (8,1.0)
    (9, 1.0)
    (10, 1.0)
    };

\addplot[thick, 
    color=red,
    mark=square, mark size=8pt
    ]
    plot coordinates {
    (6,0.89)
    };

\addlegendentry{entity-destination(e1, e2)}
\end{axis}
\end{tikzpicture}%
\caption{{\it LISA} for $S4$}\label{LISAforS4}
\end{subfigure}\hspace*{\fill}%
~%
\begin{subfigure}{0.33\textwidth}
\centering
\begin{tikzpicture}[scale=0.62,trim axis left, trim axis right]
\begin{axis}[
    ylabel={Prediction Probability},
    xmin=0, xmax=9,
    ymin=0.0, ymax=1.1,
    xtick={0,1,2,3,4,5,6,7,8},
    ytick={0,0.1,0.2,0.3,0.4,0.5,0.6,0.7, 0.8, 0.9, 1.0},
    xticklabels={,$<$e1$>$, car,  $<$/e1$>$, left, the, $<$e2$>$, plant, $<$/e2$>$},
    x tick label style={rotate=45,anchor=east},
    legend pos=south east,
    ymajorgrids=true,
    grid style=dashed,
     xlabel near ticks, compat=1.3
]
\addplot[thick, 
    color=blue,
    mark=*,
   mark options={solid}, smooth
    ]
    plot coordinates {
    (1,0.09)
    (2,0.22)
    (3,0.20)
    (4,0.33)
    (5,0.99)
    (6, 1.0)
    (7, 1.0)
    (8, 1.0)
    };
\addplot[thick, 
    color=red,
    mark=square, mark size=8pt
    ]
    plot coordinates {
   (5,0.99)
    };

\addlegendentry{entity-origin(e1, e2)}
\end{axis}
\end{tikzpicture}%
\caption{{\it LISA} for $S5$}\label{LISAforS5}
\end{subfigure}\hspace*{\fill}%
~%
\begin{subfigure}{0.33\textwidth}
\centering
\begin{tikzpicture}[scale=0.62,trim axis left, trim axis right]
\begin{axis}[
    ylabel={Prediction Probability},
    xmin=0, xmax=10,
    ymin=0.0, ymax=1.1,
    xtick={0,1,2,3,4,5,6,7,8,9},
    ytick={0,0.1,0.2,0.3,0.4,0.5,0.6,0.7, 0.8, 0.9, 1.0},
    xticklabels={,$<$e1$>$, cigarettes,  $<$/e1$>$, by, the, major, $<$e2$>$, producer, $<$/e2$>$},
    x tick label style={rotate=45,anchor=east},
    legend pos=south east,
    ymajorgrids=true,
    grid style=dashed,
     xlabel near ticks, compat=1.3
]
\addplot[thick, 
    color=blue,
    mark=*,
   mark options={solid}, smooth
    ]
    plot coordinates {
    (1,0.14)
    (2,0.14)
    (3,0.16)
    (4,0.55)
    (5,0.99)
    (6,0.99)
    (7,1.0)
    (8,1.0)
    (9,1.0)
    };

\addplot[thick, 
    color=red,
    mark=square, mark size=8pt
    ]
    plot coordinates {
    (4,0.55)
    };
\addlegendentry{product-producer(e1, e2)}
\end{axis}
\end{tikzpicture}%
\caption{{\it LISA} for $S6$}\label{LISAforS6}
\end{subfigure}

\medskip

\begin{subfigure}{0.33\linewidth}
\centering
\begin{tikzpicture}[scale=0.62,trim axis left, trim axis right]
\begin{axis}[
    ylabel={Prediction Probability},
    xmin=0, xmax=10,
    ymin=0.0, ymax=1.1,
    xtick={0,1,2,3,4,5,6,7,8,9},
    ytick={0,0.1,0.2,0.3,0.4,0.5,0.6,0.7, 0.8, 0.9, 1.0},
    xticklabels={,$<$e1$>$, cigarettes,  $<$/e1$>$, are, used, by, $<$e2$>$,  women, $<$/e2$>$},
    x tick label style={rotate=45,anchor=east},
    legend pos=south east,
    ymajorgrids=true,
    grid style=dashed,
 xlabel near ticks, compat=1.3
]
\addplot[thick, 
    color=blue,
    mark=*,
   mark options={solid}, smooth
    ]
    plot coordinates {
    (1,0.05)
    (2,0.10)
    (3,0.07)
    (4,0.30)
    (5,0.82)
    (6,0.95)
    (7,0.98)
    (8,0.99)
    (9,0.99)
    };

\addplot[thick, 
    color=red,
    mark=square, mark size=8pt
    ]
    plot coordinates {
    (5,0.82)
    };

\addlegendentry{instrument-agency(e1, e2)}
\end{axis}
\end{tikzpicture}%
\caption{{\it LISA} for $S7$}\label{LISAforS7}
\end{subfigure}\hspace*{\fill}%
~%
\begin{subfigure}{0.33\linewidth}
\centering
\begin{tikzpicture}[scale=0.62,trim axis left, trim axis right]
\begin{axis}[
    ylabel={Prediction Probability},
    xmin=0, xmax=10,
    ymin=0.0, ymax=1.1,
    xtick={0,1,2,3,4,5,6,7,8,9},
    ytick={0,0.1,0.2,0.3,0.4,0.5,0.6,0.7, 0.8, 0.9, 1.0},
    xticklabels={,$<$e1$>$, person,  $<$/e1$>$, was, born, in, $<$e2$>$,  location, $<$/e2$>$},
    x tick label style={rotate=45,anchor=east},
    legend pos=south east,
    ymajorgrids=true,
    grid style=dashed,
 xlabel near ticks, compat=1.3
]
\addplot[thick, 
    color=blue,
    mark=*,
   mark options={solid}, smooth
    ]
    plot coordinates {
    (1,0.34)
    (2,0.34)
    (3,0.34)
    (4,0.37)
    (5,0.50)
    (6,0.58)
    (7,0.53)
    (8,0.54)
    (9,0.53)
    };

\addplot[thick, 
    color=red,
    mark=square, mark size=8pt
    ]
    plot coordinates {
    (6,0.58)
    };

\addlegendentry{{\it slot} per:location\_of\_birth}
\end{axis}
\end{tikzpicture}%
\caption{{\it LISA} for $S8$}\label{LISAforS8}
\end{subfigure}\hspace*{\fill}%
~%
\begin{subfigure}{0.33\textwidth}
\centering
\begin{tikzpicture}[scale=0.62,trim axis left, trim axis right]
\begin{axis}[
    ylabel={Prediction Probability},
    xmin=0, xmax=8,
    ymin=0.0, ymax=1.1,
    xtick={0,1,2,3,4,5,6,7},
    ytick={0,0.1,0.2,0.3,0.4,0.5,0.6,0.7, 0.8, 0.9, 1.0},
    xticklabels={,$<$e1$>$, person,  $<$/e1$>$, married, $<$e2$>$, spouse, $<$/e2$>$},
    x tick label style={rotate=45,anchor=east},
    legend pos=south east,
    ymajorgrids=true,
    grid style=dashed,
    xlabel near ticks, compat=1.3
]
\addplot[thick, 
    color=blue,
    mark=*,
   mark options={solid}, smooth
    ]
    plot coordinates {
    (1,0.02)
    (2,0.02)
    (3,0.10)
    (4,0.98)
    (5,0.99)
    (6, 1.0)
    (7,0.98)
    };

\addplot[thick, 
    color=red,
    mark=square, mark size=8pt
    ]
    plot coordinates {
      (4,0.98)
    };

\addlegendentry{{\it slot} per:spouse}
\end{axis}
\end{tikzpicture}%
\caption{{\it LISA} for $S9$}\label{LISAforS9}
\end{subfigure}

\medskip

\begin{subfigure}{0.48\textwidth} 
		\includegraphics[height=6cm, width=\textwidth]{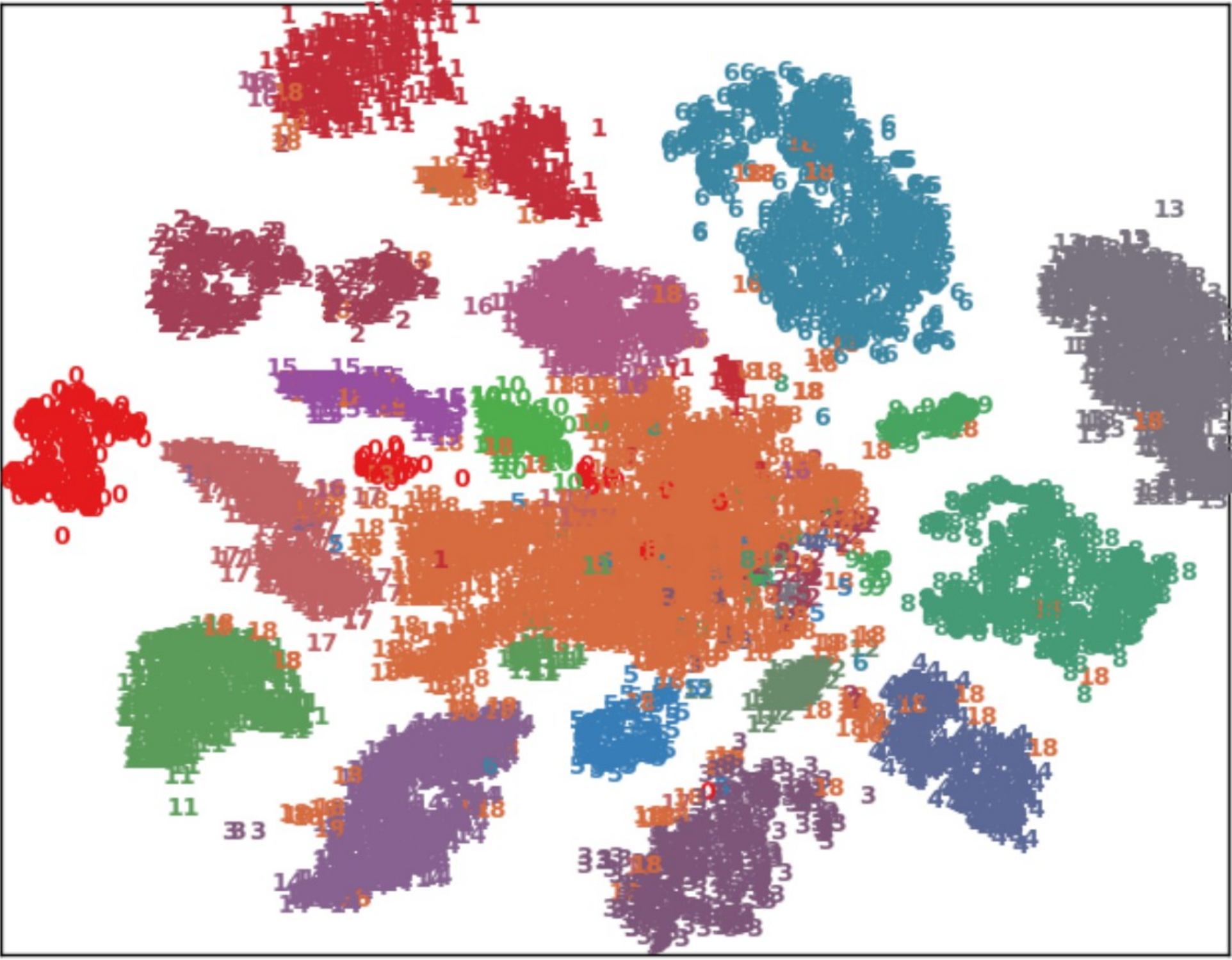}
		\caption{t-SNE Visualization for training set}
                      \label{tsne18classestrainingdata}
\end{subfigure}
\hspace{0.5cm}
\begin{subfigure}{0.48\textwidth} 
		\includegraphics[height=6cm, width=\textwidth]{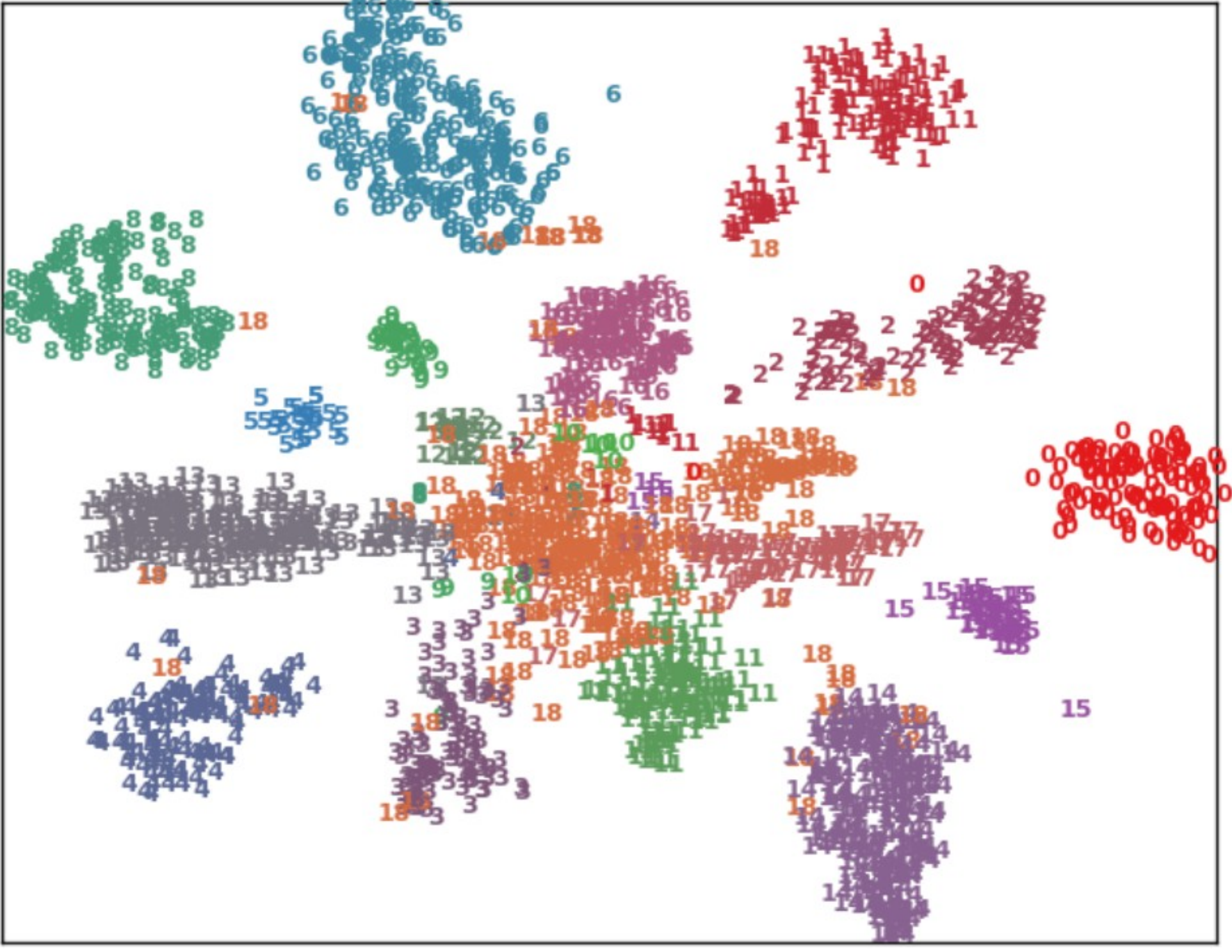}
		\caption{t-SNE Visualization for testing set}
                      \label{tsne18classestestingdata}
\end{subfigure}
\caption{(a-i) Layer-wIse Semantic Accumulation (LISA) by C-BRNN for different relation types in SemEval10 Task 8 and TAC KBP Slot Filling datasets. 
The square in red color signifies that the relation is correctly detected with the input subsequence (enough in decision making).  
(j-k) t-SNE visualization of the last combined hidden unit ($h_{bi}$) of C-BRNN computed using the SemEval10 train and test sets.}
\end{figure*}

\begin{table*}[t]
\centering
\def\arraystretch{1.1}
\resizebox{0.99\textwidth}{!}{%
\begin{tabular}{c|l|l|l}
\multicolumn{1}{c|}{\bf ID} & \multicolumn{1}{c|}{\bf Relation/Slot Types}  & \multicolumn{1}{c|}{\bf Example Sentences} & \multicolumn{1}{c}{\bf Example2Pattern}\\ \hline
$S1$   &  cause-effect(e1, e2)        & $<$e1$>$ demolition $<$/e1$>$ was the cause of $<$e2$>$ terror $<$/e2$>$ & cause of  $<$e2$>$ \\
$S2$  &  cause-effect(e2, e1)        & $<$e1$>$ damage $<$/e1$>$ caused by the $<$e2$>$ bombing $<$/e2$>$    & damage $<$/e1$>$ caused\\
$S3$ & component-whole(e1, e2)   & $<$e1$>$ countyard $<$/e1$>$ of the $<$e2$>$ castle $<$/e2$>$      &   $<$/e1$>$ of the\\
$S4$  &  entity-destination(e1,e2)  & $<$e1$>$  marble  $<$/e1$>$ was dropped into the $<$e2$>$   bowl  $<$/e2$>$   &  dropped into the\\
$S5$ & entity-origin(e1, e2)  & $<$e1$>$ car $<$/e1$>$ left the $<$e2$>$ plant $<$/e2$>$  & left the $<$e2$>$\\
$S6$ & product-produce(e1, e2)  & $<$e1$>$ cigarettes $<$/e1$>$ by the major $<$e2$>$ producer $<$/e2$>$  & $<$/e1$>$ by the\\
$S7$  & instrument-agency(e1, e2)   & $<$e1$>$ cigarettes $<$/e1$>$ are used by $<$e2$>$ women $<$/e2$>$    &   $<$/e1$>$ are used \\
$S8$  & per:loc\_of\_birth(e1, e2) & $<$e1$>$  person $<$/e1$>$  was born in $<$e2$>$ location  $<$/e2$>$  & born in $<$e2$>$\\ 
$S9$ & per:spouse(e1, e2)   & $<$e1$>$  person $<$/e1$>$  married $<$e2$>$ spouse $<$/e2$>$  & $<$/e1$>$ married $<$e2$>$\\ \hline
\end{tabular}}
\caption{Example Sentences for {\it LISA} and {\it example2pattern} illustrations. The sentences $S1$-$S7$ belong to SemEval10 Task 8 dataset and $S8$-$S9$ to TAC KBP Slot Filling (SF) shared task dataset.} 
\label{semeval10sentences}
\end{table*}

{\bf LISA Formulation}: To explain the cumulative nature of recurrent neural networks, we show how does it build semantic meaning of a sentence  
word by word  belonging to a particular category in the data and compute prediction scores for the expected category on different inputs, as shown in Figure \ref{fig:scorecomputing}. The scheme also depicts the contribution of each word in the sequence towards the final classification score (prediction probability).   

At first, we compute different subsequences of word(s) for a given sequence of words (i.e., sentence).  
Consider a sequence ${\bf S}$ of words [$w_1, w_2, ..., w_k, ..., w_n$] for a given sentence $S$ 
of length $n$. We compute $n$ number of subsequences, where each subsequence ${\bf S}_{\le k}$ is a 
subvector of words [$w_1, ...w_k$], i.e.,  ${\bf S}_{\le k}$ consists of words preceding and including the word $w_k$ 
in the sequence ${\bf S}$. In context of this work, extending a subsequence by a word means 
appending the subsequence by the next word in the sequence.   
Observe that the number of subsequences, $n$ is equal to the total number of time steps in the C-BRNN.   

Next is to compute RNN prediction score for the category $R$ associated with sentence $S$. 
We compute the score via the autoregressive conditional $P(R|{\bf S}_{\le k}, \mathbb{M})$ for each subsequence ${\bf S}_{\le k}$, as-
\begin{equation}\label{LISA}
P(R|{\bf S}_{\le k}, \mathbb{M}) = softmax (W_{hy} \cdot h_{{bi}_{k}} + b_y)
\end{equation}

using the trained C-BRNN (Figure \ref{fig:CBRNN}) model $\mathbb{M}$. For each $k \in [1, n]$, 
we compute the network prediction, $P(R|{\bf S}_{\le k}, \mathbb{M})$ to demonstrate the cumulative property of recurrent neural network that 
builds meaningful semantics of the sequence $\bf S$ by extending each subsequence ${\bf S}_{\le k}$ word by word.    
The internal state $h_{{bi}_{k}}$ (attached to softmax layer as in Figure \ref{fig:CBRNN}) is involved in decision making for each  input subsequence ${\bf S}_{\le k}$ with bias vector $b_y \in \mathbb{R}^{C}$ and hidden-to-softmax weights matrix $W_{hy} \in \mathbb{R}^{D \times C}$ for $C$ categories.

The {\it LISA} is illustrated in Figure \ref{fig:scorecomputing}, where each word in the sequence contributes to final classification score. 
It allows us to understand the network decisions via peaks  in the prediction score over different subsequences.   
The peaks signify the saliency patterns (i.e., sequence of words) that the network has learned in order to make decision.  
For instance, the input word `{\it of}' following the subsequence `{\it $<$e1$>$ demolition $<$/e1$>$  was the cause}' introduces a sudden increase in  prediction score for  the relation type {\it cause}-{\it effect}(e1, e2). 
It suggests that the C-BRNN collects the semantics layer-wise via temporally organized subsequences.
Observe that the subsequence `{\it ...cause of}' is salient enough in decision making (i.e., prediction score=$0.77$), 
where the next subsequence `{\it ...cause of $<$e2$>$}'  adds in the score to get $0.98$.

\begin{algorithm}[t]
\caption{Example2pattern Transformation}\label{example2pattern}
\begin{algorithmic}[1]
\Statex {\bf Input:} sentence $S$, length $n$, category $R$, threshold $\tau$, C-BRNN $\mathbb{M}$, N-gram size N
\Statex {\bf Output:} N-gram saliency pattern $patt$
\For{$k$ in $1$ to $n$}
\State {compute} $\mbox{N-gram}_k$  (eqn \ref{ngram}) of words in $S$ 
\EndFor
\For{$k$ in $1$ to $n$}
       \State {compute} ${\bf S}_{\le k}$ (eqn \ref{trigramsequence2}) of N-grams
       \State {compute} $P(R|{\bf S}_{\le k}, \mathbb{M})$  using eqn \ref{LISA}
        \If {$ P(R|{\bf S}_{\le k}, \mathbb{M}) \ge \tau $} \\ 
           \qquad  \qquad \Return $patt \gets {\bf S}_{\le k}[-1]$
           \EndIf
\EndFor
\end{algorithmic}
\end{algorithm}

{\bf Example2pattern for Saliency Pattern}: 
To further interpret RNN, we seek to identify and extract the most likely input pattern (or phrases) for a given class that is discriminating enough in decision making. 
Therefore, each example input is transformed into a saliency pattern that informs us about the network learning.    
To do so, we first compute N-gram for each word $w_t$ in the sentence $S$. 
For instance, a 3-gram representation of $w_t$ is given by $w_{t-1}, w_t, w_{t+1}$.  
Therefore, an N-gram (for N=3) sequence $\bf S$ of words is represented as  $[[w_{t-1}, w_t, w_{t+1}]_{t=1}^{n}]$, where $w_0$ and $w_{n+1}$ are PADDING (zero) vectors of embedding dimension. 

Following \newcite{Thang:82}, we use N-grams (e.g., tri-grams) representation for each word in each subsequence ${\bf S}_{\le k}$ that is input to C-BRNN to compute $P(R | {\bf S}_{\le k})$, where the N-gram (N=3) subsequence ${\bf S}_{\le k}$ is given by, 
\begin{align}
\begin{split}\label{trigramsequence1}
{\bf S}_{\le k} & = [ [PADDING, w_1, w_2]_1,  [w_1, w_2, w_3]_2,...,\\ & [w_{t-1}, w_{t}, w_{t+1}]_t, ..., [w_{k-1}, w_{k}, w_{k+1}]_{k}]
\end{split}\\
\begin{split}\label{trigramsequence2}
& {\bf S}_{\le k} =  [tri_1, tri_2, ..., tri_t, ...tri_{k}]
\end{split}
\end{align}
for $k \in [1, n]$. Observe that the 3-gram $tri_k$ consists of the word $w_{k+1}$, if k $\neq$ n. 
To generalize for $i \in [1, \floor{N/2}]$, an $\mbox{N-gram}_k$ of size $N$ for word $w_k$ in C-BRNN is given by- 
\begin{equation}\label{ngram}
\mbox{N-gram}_k = [w_{k-i}, ...,  w_{k}, ..., w_{k+i}]_k 
\end{equation}

Algorithm \ref{example2pattern} shows the transformation of an example sentence into pattern that is salient in decision making.   
For a given example sentence $S$ with its length $n$ and category $R$, we extract the most salient N-gram (N=3, 5 or 7) 
pattern $patt$ (the last N-gram in the N-gram subsequence ${\bf S}_{\le k}$) that contributes the most in detecting the relation type $R$. 
The threshold parameter $\tau$ signifies the probability of prediction for the category $R$ by the model $\mathbb{M}$. 
For an input N-gram sequence ${\bf S}_{\le k}$ of sentence $S$, we extract the last N-gram, e.g., $tri_{k}$  that detects the relation $R$ with prediction 
score above $\tau$.  By manual inspection of patterns extracted at different values (0.4, 0.5, 0.6, 0.7) of $\tau$, 
we found that $\tau=0.5$ generates the most salient and interpretable patterns.  
The saliency pattern detection follows LISA as demonstrated in Figure \ref{fig:scorecomputing}, 
except that we use N-gram ($N=$3, 5 or 7) input to detect and extract the key relationship patterns. 

\begin{table}[t]
\centering
\def\arraystretch{1.1}
\resizebox{0.49\textwidth}{!}{%
\begin{tabular}{l|c}
\multicolumn{1}{c|}{\bf Input word sequence to C-BRNN} & \multicolumn{1}{c}{$pp$}\\ \hline
 $\bf <${\bf e1}$\bf >$ &  $0.10$  \\
$<$e1$>$   {\bf demolition} & $ 0.25$\\
$<$e1$>$   demolition  $\bf <${\bf /e1}$\bf >$ & $0.29$\\
$<$e1$>$   demolition $<$/e1$>$ {\bf was} & $0.30$\\
$<$e1$>$   demolition $<$/e1$>$ was {\bf the} &  $0.35$\\
$<$e1$>$   demolition $<$/e1$>$ was the {\bf cause} & $0.39$\\
$<$e1$>$   demolition $<$/e1$>$ was the cause {\bf of}  &  \underline{$0.77$}\\
$<$e1$>$   demolition $<$/e1$>$ was the cause of  $\bf <${\bf e2}$\bf >$ & $0.98$\\
$<$e1$>$   demolition $<$/e1$>$ was the cause of $<$e2$>$ {\bf terror} & $1.00$\\
$<$e1$>$   demolition $<$/e1$>$ was the cause of $<$e2$>$ terror $\bf <${\bf /e2}$\bf >$  &    $1.00$ \\ \hline
\end{tabular}}
\caption{Semantic accumulation and sensitivity of C-BRNN over subsequences for sentence $S1$.  
Bold indicates the last word in the subsequence. $pp$: prediction probability in the softmax layer for the relation type. 
The underline signifies that the $pp$ is sufficient enough ($\tau$=$0.50$) in detecting the relation.  
Saliency patterns, i.e., N-grams can be extracted from the input subsequence that leads to a sudden peak in $pp$, where $pp \ge \tau$.}
\label{semanticaccwordinput}
\end{table}

\begin{table*}[t]
\centering
\def\arraystretch{1.1}
\resizebox{0.99\textwidth}{!}{%
\begin{tabular}{l|c|c|c}
\multicolumn{1}{c|}{\bf Relation} &  \multicolumn{1}{c|}{\bf 3-gram Patterns}  &  \multicolumn{1}{c|}{\bf 5-gram Patterns}  &  \multicolumn{1}{c}{\bf 7-gram Patterns}\\ \hline
 &  $<$/e1$>$ cause $<$e2$>$          &    the leading causes of $<$e2$>$    &  is one of the leading causes of   \\ 
{\it cause}- &  $<$/e1$>$ caused a            &    the main causes of $<$e2$>$    &    is one of the main causes of  \\
{\it effect}(e1,e2)  &   that cause respiratory           &    $<$/e1$>$ leads to $<$e2$>$ inspiration   &    $<$/e1$>$ that results in $<$e2$>$ hardening $<$/e2$>$  \\
&   which cause acne            &    $<$/e1$>$ that results in $<$e2$>$   &     $<$/e1$>$ resulted in the $<$e2$>$ loss $<$/e2$>$ \\
&  leading causes of    &   $<$/e1$>$ resulted in the $<$e2$>$   &  $<$e1$>$ sadness $<$/e1$>$ leads to $<$e2$>$ inspiration\\ \hline

&   caused due to                                        &      $<$/e1$>$ has been caused by                                 &      $<$/e1$>$ is caused by a $<$e2$>$ comet                       \\
&    comes from the                                          &    $<$/e1$>$ are caused by the                                       &   $<$/e1$>$ however has been caused by the \\
{\it cause}- &   arose from an                     &     $<$/e1$>$ arose from an $<$e2$>$         &          $<$/e1$>$ that has been caused by the \\
{\it effect}(e2,e1) &   caused by the                 &         $<$/e1$>$ caused due to $<$e2$>$           &       that has been caused by the $<$e2$>$  \\
 &   radiated from a                                &      infection $<$/e2$>$ results in an            &            $<$e1$>$ product $<$/e1$>$ arose from an $<$e2$>$      \\ \hline 

&  in a $<$e2$>$               &  $<$/e1$>$ was contained in a                                       &               $<$/e1$>$ was contained in a $<$e2$>$ box             \\  
&   was inside a            &              $<$/e1$>$ was discovered inside a                        &        $<$/e1$>$ was in a $<$e2$>$ suitcase $<$/e2$>$                  \\
{\it content}- &   contained in a                     &         $<$/e1$>$ were in a $<$e2$>$                             &              $<$/e1$>$ were in a $<$e2$>$ box $<$/e2$>$               \\
{\it container}(e1,e2)  &  hidden in a                      &            is hidden in a $<$e2$>$                           &        $<$/e1$>$ was inside a $<$e2$>$ box $<$/e2$>$                     \\
 &  stored in a                      &                 $<$/e1$>$ was contained in a                     &            $<$/e1$>$ was hidden in an $<$e2$>$ envelope                  \\ \hline

& $<$/e1$>$ released by     &         $<$/e1$>$ issued by the $<$e2$>$                                         &          $<$e1$>$ products $<$/e1$>$ created by an $<$e2$>$            \\
{\it product}-  &  $<$/e1$>$ issued by       &     $<$/e1$>$ was prepared by $<$e2$>$                                             &       $<$/e1$>$ by an $<$e2$>$ artist $<$/e2$>$ who               \\
{\it produce}(e1,e2) &  $<$/e1$>$ created by                     &    was written by a $<$e2$>$                                          &             $<$/e1$>$ written by most of the $<$e2$>$         \\ 
&  by the $<$e2$>$   &         $<$/e1$>$ built by the $<$e2$>$                                                               &              temple $<$/e1$>$ has been built by $<$e2$>$         \\
&  of the $<$e1$>$            &      $<$/e1$>$ are made by $<$e2$>$                                             &                 $<$/e1$>$ were founded by the $<$e2$>$ potter      \\ \hline

& $<$/e1$>$  of the                              &          $<$/e1$>$ of the $<$e2$>$ device                                                   &        the $<$e1$>$ timer $<$/e1$>$ of the $<$e2$>$                                            \\
whole(e1, e2) &  of the $<$e2$>$                               &     $<$/e1$>$ was a part of                                                         &                 $<$/e1$>$ was a part of the romulan   \\
component- &  part of the                               &       $<$/e1$>$ is part of the                                                         &                $<$/e1$>$ was the best part of the                               \\
&   $<$/e1$>$ of $<$e2$>$                               &          is a basic element of                                                  &    $<$/e1$>$ is a basic element of the                                                  \\
&  $<$/e1$>$ on a                                                 & $<$/e1$>$ is part of a     &                  are core components of the $<$e2$>$ solutions                                    \\ \hline

&  put into a                &                         have been moving into the                 &                 $<$/e1$>$ have been moving back into $<$e2$>$          \\ 
&  released into the                 &            was dropped into the $<$e2$>$                              &             $<$/e1$>$ have been moving into the $<$e2$>$               \\ 
entity-& $<$/e1$>$ into the                 &                 $<$/e1$>$ moved into the $<$e2$>$                         &        $<$/e1$>$ have been dropped into the $<$e2$>$                   \\ 
destination(e1,e2) &  moved into the                 &         were released into the $<$e2$>$                                 &            $<$/e1$>$ have been released back into the                \\ 
& added to the                 &         $<$/e1$>$ have been exported to                                 &         power $<$/e1$>$ is exported to the $<$e2$>$                   \\  \hline

 &  $<$/e1$>$ are used         &               $<$/e1$>$ assists the $<$e2$>$ eye                             &               cigarettes $<$/e1$>$ are used by $<$e2$>$ women             \\
 &  used by $<$e2$>$         &              $<$/e1$>$ are used by $<$e2$>$                              &               $<$e1$>$ telescope $<$/e1$>$ assists the $<$e2$>$ eye           \\
 instrument-&  $<$/e1$>$ is used         &           $<$/e1$>$ were used by some                                 &        $<$e1$>$ practices $<$/e1$>$ for $<$e2$>$ engineers $<$/e2$>$                    \\
 agency(e1,e2)&  set by the         &           $<$/e1$>$ with which the $<$e2$>$                                 &                the best $<$e1$>$ tools $<$/e1$>$ for $<$e2$>$            \\
 &  $<$/e1$>$ set by         &                   readily associated with the $<$e2$>$                         &                    $<$e1$>$ wire $<$/e1$>$ with which the $<$e2$>$        \\ \hline

\end{tabular}}
\caption{SemEval10 Task 8 dataset: N-Gram (3, 5 and 7) saliency patterns extracted for different relation types by C-BRNN with PI}
\label{n-gramSemEval10}
\end{table*}

\section{Analysis: Relation Classification}
Given a sentence and two annotated nominals, the task of binary relation classification is to predict the semantic relations between the pairs of nominals.  
In most cases, the context in between the two nominals define the relationship.  However, \newcite{Thang:82} has shown that the extended context helps. 
In this work, we focus on the building semantics for a given sentence using relationship contexts between the two nominals. 

We analyse RNNs for {\it LISA} and {\it example2pattern} using two relation classification datsets:   
(1) SemEval10 Shared Task 8 \cite{hendrickx2009semeval} (2) TAC KBP Slot Filling (SF) shared task\footnote{data from the slot filler classification component of the slot filling pipeline, treated as relation classification}  
\cite{adel2015cis}. 
We demonstrate the sensitiveness of RNN for different subsequences (Figure \ref{fig:scorecomputing}), input in the same order as in the original sentence. 
We explain its predictions (or judgments) and extract the salient relationship patterns learned for each category in the two datasets.  

\subsection{SemEval10 Shared Task 8 dataset}\label{semeval10analysis}
 The relation classification dataset of the Semantic Evaluation 2010 (SemEval10) shared task 8 \cite{hendrickx2009semeval} consists of 
19 relations (9 directed relations and one artificial class \texttt{Other}), 8,000 training  and 2,717 testing sentences. 
We split the training data into train (6.5k) and development (1.5k) sentences  to optimize the C-BRNN network. 
For instance, an example sentence with relation label is given by-

\texttt{The $<$e1$>$ demolition $<$/e1$>$ was the cause of $<$e2$>$ terror $<$/e2$> $ and communal divide is just a way of not letting truth prevail. $\rightarrow$ cause-effect(e1,e2)}

The terms  \texttt{demolition} and \texttt{terror} are the relation arguments or nominals, where the phrase \texttt{was the cause of} is the relationship context between the two arguments. 
Table \ref{semeval10sentences} shows the examples sentences (shortened to argument1+relationship context+argument2) drawn from the development and test sets that we employed 
to analyse the C-BRNN for semantic accumulation in our experiments.  
We use the similar experimental setup as \newcite{Thang:82}.

\begin{table}[t]
\centering
\def\arraystretch{1.05}
\resizebox{0.44\textwidth}{!}{%
\begin{tabular}{l|c}
\multicolumn{1}{c|}{\bf Slots} &  \multicolumn{1}{c}{\bf N-gram Patterns}\\ \hline
&  $<$/e1$>$ wife of\\
 & $<$/e1$>$ , wife\\
{\it per}- &   $<$/e1$>$ ’ wife\\
{\it spouse}(e1,e2) &  $<$/e1$>$ married $<$e2$>$\\
&  $<$/e1$>$ marriages to\\ \hline

&     was born in\\
& born in $<$e2$>$\\
{\it per}-& a native of\\
{\it location\_of\_birth}(e1,e2) &  $<$/e1$>$ from $<$e2$>$\\
&  $<$/e1$>$ 's  hometown\\ \hline
\end{tabular}}
\caption{TAC KBP SF dataset: Tri-gram saliency patterns extracted for slots {\it per}:{\it spouse}(e1, e2) and {\it per}:{\it location\_of\_birth}(e1,e2)}
\label{n-gramTACKBP}
\end{table}

{\it LISA Analysis}: As discussed in Section \ref{sec:LISAexample2pattern}, we interpret C-BRNN by explaining its predictions via the semantic accumulation over the subsequences ${\bf S}_{\le k}$ (Figure \ref{fig:scorecomputing}) for each sentence $S$.    
We select the example sentences $S1$-$S7$ (Table \ref{semeval10sentences}) for which the network predicts the correct relation type with high scores.  
For an example sentence $S1$, Table \ref{semanticaccwordinput} illustrates how different subsequences are input to C-BRNN in order to compute prediction scores $pp$ 
in the softmax layer for the relation \texttt{cause-effect(e1, e2)}. We use tri-gram (section \ref{sec:LISAexample2pattern}) word representation for each word for the examples $S1$-$S7$. 

Figures \ref{LISAforS1},  \ref{LISAforS2},  \ref{LISAforS3},  \ref{LISAforS4}  \ref{LISAforS5},  \ref{LISAforS6} and  \ref{LISAforS7} demonstrate the cumulative nature 
and sensitiveness of RNN via prediction probability ($pp$) about different inputs for sentences $S1$-$S7$, respectively.  
For instance in Figure \ref{LISAforS1} and Table \ref{semanticaccwordinput}, the C-BRNN builds meaning of the sentence $S1$ word by word, where a sudden increase in $pp$ is observed 
when the input subsequence \texttt{$<$e1$>$ demolition $<$/e1$>$ was the cause} is extended with the next term \texttt{of} in the word sequence ${\bf S}$.  
Note that the relationship context between the arguments \texttt{demolition} and \texttt{terror} is sufficient enough in detecting the relationship type.  
Interestingly, we also observe that the prepositions (such as \texttt{of}, \texttt{by}, \texttt{into}, etc.) in combination with verbs are key features in building the meaningful semantics.    


{\it Saliency Patterns via example2pattern Transformation}:
Following the discussion in Section \ref{sec:LISAexample2pattern} and Algorithm \ref{example2pattern}, we transform each correctly identified example into pattern by 
extracting the most likely N-gram in the input subsequence(s).  
In each of the Figures \ref{LISAforS1},  \ref{LISAforS2},  \ref{LISAforS3},  \ref{LISAforS4}  \ref{LISAforS5},  \ref{LISAforS6} and  \ref{LISAforS7}, the square box in red color 
signifies that the relation type is correctly identified (when $\tau =0.5$)  at this particular subsequence input (without the remaining context in the sentence). 
We extract the last N-gram of such a subsequence.   

Table \ref{semeval10sentences} shows the {\it example2pattern} transformations for sentences $S1$-$S7$ in SemEval10 dataset, derived from 
Figures \ref{LISAforS1}-\ref{LISAforS7}, respectively with N=3 (in the N-grams).  
Similarly, we extract the salient patterns (3-gram, 5-gram and 7-gram) (Table \ref{n-gramSemEval10}) for different relationships.   
We also observe that the relation types \texttt{content-container(e1, e2)} and  \texttt{instrument-agency(e1, e2)} are mostly defined by smaller relationship contexts (e.g, 3-gram), however \texttt{entity-destination(e1,e2)}  by larger contexts (7-gram). 


\subsection{TAC KBP Slot Filling dataset}
We investigate another dataset from TAC KBP Slot Filling (SF)  shared task \cite{surdeanu2013overview}, where we use the relation classification dataset by \newcite{HeikeslotfillingCNNs2016}
 in the context of slot filling. We have selected the two slots: \textit{per:loc\_of\_birth} and \textit{per:spouse} out of 24 types.

{\it LISA Analysis}: Following Section \ref{semeval10analysis}, we analyse the C-BRNN for LISA using sentences $S8$ and $S9$ (Table \ref{semeval10sentences}). 
Figures \ref{LISAforS8} and \ref{LISAforS9} demonstrate the cumulative nature of recurrent neural network, 
where we observe that the salient patterns \texttt{born in $<$e2$>$} and \texttt{$<$/e1$>$ married $e2$} lead to correct decision making for $S8$ and $S9$, respectively. 
Interestingly for $S8$, we see a decrease in prediction score from $0.59$  to $0.52$ on including terms in the subsequence, following the term \texttt{in}.  


{\it Saliency Patterns via example2pattern Transformation}: 
Following  Section \ref{sec:LISAexample2pattern} and Algorithm \ref{example2pattern}, we demonstrate the {\it example2pattern} transformation of sentences $S8$ and $S9$ in Table \ref{semeval10sentences} with 
tir-grams. In addition, Table \ref{n-gramTACKBP} shows the tri-gram salient patterns extracted for the two slots.

\section{Visualizing Latent Semantics}
In this section, we attempt to visualize the hidden state of each test (and train) example that has accumulated (or built) the meaningful semantics during sequential processing in C-BRNN.
To do this, we compute the last hidden vector $h_{bi}$ of the combined network (e.g., $h_{bi}$ attached to the softmax layer in Figure \ref{fig:CBRNN}) for each test (and train) example and 
visualize (Figure \ref{tsne18classestestingdata} and \ref{tsne18classestrainingdata}) using t-SNE \cite{maaten2008visualizing}. 
Each color represents a relation-type.  Observe the distinctive
clusters of accumulated semantics in hidden states for each category in the data (SemEval10 Task 8).


\section{Conclusion and Future Work}
We have demonstrated the cumulative nature of recurrent neural networks via sensitivity analysis over different inputs, i.e., {\it LISA} 
to understand how they build meaningful semantics and explain predictions for each category in the data. We have also detected a salient pattern in each of the example sentences, i.e., {\it example2pattern transformation} 
that the network learns in decision making. We extract the salient patterns for different categories in two relation classification datasets.

In future work, it would be interesting to analyse the sensitiveness of RNNs with corruption in the salient patterns.  
One could also investigate visualizing the dimensions of hidden states (activation maximization) and word embedding vectors with the network decisions over time. 
We forsee to apply {\it LISA} and {\it example2pattern} on different tasks such as document categorization, sentiment analysis, language modeling, etc.  Another interesting direction would be to analyze the bag-of-word neural topic models 
such as DocNADE  \cite{docnadehugo:82} and iDocNADE \cite{GuptaiDocNADE:85} to interpret their semantic accumulation during autoregressive computations in  building document representation(s).      
We extract the saliency patterns for each category in the data 
that can be 
effectively used in instantiating pattern-based information extraction systems, such as bootstrapping entity \cite{gupta2014spied} and relation extractors \cite{GuptabootRE:82}. 


\section*{Acknowledgments}
We thank Heike Adel for providing us with the TAC KBP dataset used in our experiments. 
We express appreciation for our colleagues  Bernt Andrassy, Florian Buettner,  Ulli Waltinger, Mark Buckley,  Stefan Langer, Subbu Rajaram, Yatin Chaudhary, and 
anonymous reviewers for their in-depth review comments. 
This research was supported by Bundeswirtschaftsministerium ({\tt bmwi.de}), grant 01MD15010A (Smart Data Web) 
at Siemens AG- CT Machine Intelligence, Munich Germany. 

\bibliography{emnlp2018}
\bibliographystyle{emnlp2018}

\end{document}